\pdfoutput=1
\documentclass[sigconf]{IEEEtran}
\usepackage{mathrsfs}
\usepackage{cite}
\usepackage{amsmath,amssymb,amsfonts}
\usepackage{graphicx}
\usepackage{textcomp}
\usepackage{xcolor}
\usepackage{pgfplots}
\pgfplotsset{compat=1.14}

\begin{document}

\title{Micro-Browsing Models for Search Snippets}

\author{Muhammad Asiful Islam, Ramakrishnan Srikant, Sugato Basu}

\author{\IEEEauthorblockN{Muhammad Asiful Islam, Ramakrishnan Srikant, Sugato Basu}
\IEEEauthorblockA{ \\
\textit{Google Inc}\\
Mountain View, CA, USA \\
maislam,srikant,sugato@google.com}
}

\maketitle

\begin{abstract}
Click-through rate (CTR) is a key signal of relevance for search engine results,
both organic and sponsored.  CTR of a result has two core components: (a) the
probability of examination of a result by a user, and (b) the perceived
relevance of the result given that it has been examined by the user.
There has been considerable work on user browsing models, to model and analyze
both the examination and the relevance components of CTR. In this paper,
we propose a novel formulation: a micro-browsing model for how users read result
snippets. The snippet text of a result often plays a critical role in the perceived
relevance of the result. We study how particular words {\em within} a line of
snippet can influence user behavior. We validate this new micro-browsing user
model by considering the problem of predicting which snippet will yield higher
CTR, and show that classification accuracy is dramatically higher with our
micro-browsing user model. The key insight in this paper is that varying
relatively few words within a snippet, and even their location within a snippet,
can have a significant influence on the clickthrough of a snippet.
\end{abstract}

\begin{IEEEkeywords}
Sponsored Search, Machine Learning
\end{IEEEkeywords}

\section{Introduction}
\label{sec:introduction}

Web search engines have become an essential tool for navigating the vast amounts of
information on the internet.  Implicit user feedback, specifically, click-through
data, is valuable for optimizing search engine results
\cite{agichtein06:improve_web_search, joachims02:optimize_search_engines,
radlinski05:query_chains, chen12:noise_aware}.
Click-through data plays an equally important role in estimating the quality of
sponsored search results \cite{predict:richardson}.

When using click-through data, the
position bias of a result has to be always taken into account, since
the same result usually gets a higher click-through rate (CTR) if it
is positioned in a more prominent part (e.g., the top) of the page.
There have been many papers on better estimating the probability of
examination of a result, by using the pattern of clicks on prior
results \cite{dcm:craswell, dupret:piwowarski, ccm:guoliukannan,
liu:bbm}, or using both prior clicks and the relevance of prior
results
\cite{chapelle09:bayesian_click_model, dcm:guoliuwang, zhu:gcm, srikant:rve}.
These models are all {\em macro-models}, as they build
models of examining the complete result without looking at features
based on the content of the result (e.g., snippets of organic search
results or creatives of sponsored search results).

However, the result snippet often contains certain words or phrases that make
the user more likely to click the result.  For example, a user looking for
spacious seats while booking a flight might consider ``more legroom'' to be the
most salient feature in the decision to click, whereas another user looking for
a better deal may be interested in phrases like ``cheap flights'' or ``20\%
off''.  Once the user sees these words in the snippet, she may decide to click
without examining the other words.

\vspace{1ex}
\noindent{\bf Contributions} \hspace{0.5ex}
The key insight in this paper is that relatively few word variations within
a snippet, and even where within a snippet particular words are located,
can have a meaningful influence on the clickthrough of a snippet.
Section~\ref{sec:related}
discusses related research. In Section~\ref{sec:model}, we propose a
fine-grained model for estimating the
probability that users actually read a word (or phrase) in the snippet.  We call
this the {\em micro-browsing model} for search result snippets.
Section~\ref{sec:technical} discusses the application of the
micro-browsing model to the problem of predicting which of two creatives
will have higher CTR.  Detailed experimental results and analysis are
presented in Sections~\ref{sec:experiments}, to demonstrate the
effectiveness of the micro-browsing model.
Finally Section~\ref{sec:conclusions} outlines some promising areas of future work.

\section{Related Work}
\label{sec:related}

Prior user browsing models for web search results can be partitioned into three
groups based on how they estimate the probability that the user examines a
specific position:
\begin{itemize}
\item Models that assume examination is independent of the other results for the query.
\item Models that assume examination depends on the pattern of clicks on prior results.
\item Models that assume examination depends on both the pattern of clicks on prior
  results, and the relevance of prior results.
\end{itemize}
Some of the models were originally targeted at sponsored search, while others were
targeted at organic search results.  However, while the parameter values might
differ, all of these models are general enough to apply to both organic search
and sponsored search.

\subsection{Examination independent of other results}
\label{sec:background-independent}

We use the following notation:
\begin{itemize}
\item Let $\phi(i)$ denote the result at position $i$ in a ranked list of
  results (whether organic results or sponsored results).
\item Let $C_i$ denote a binary random variable that
captures the event that a user clicks on $\phi(i)$.
\item Let $E_i$ denote the event that the user examines $\phi(i)$.
\end{itemize}

The {\bf examination hypothesis}, originally proposed by Richardson et
al. \cite{predict:richardson} and formalized by Craswell et
al. \cite{dcm:craswell}, observes that to be clicked, a result must be both
examined and relevant:
\begin{equation}
\Pr(C_i = 1) = \Pr(C_i = 1|E_i = 1) \Pr(E_i = 1).
\label{eqn:examination-hypothesis}
\end{equation}
Richardson et al. \cite{predict:richardson} assume that the probability a result
is viewed depends solely on its position, and is independent of other results.

\subsection{Examination depends on prior clicks}
\label{sec:background-clicks}

An implicit assumption in the above formulation is that the
probability of examining the result in position $i$ does not depend on
click events in other result positions.  A plethora of papers explore
models that incorporate this information into the examination probabilities.

The {\bf cascade hypothesis} \cite{dcm:craswell} assumes that users scan each
result sequentially without any skips:
\begin{align*}
\Pr(E_1 = 1) = 1, \\
\Pr(E_i = 1 | E_{i-1} = 0) =  0.
\end{align*}

The {\bf cascade model} \cite{dcm:craswell} further constrains that the user
continues examining results until she clicks on a result, and does not examine any
additional results after the click:
\begin{equation}
\Pr(E_i = 1 | E_{i-1} =  1) = 1 - C_{i-1}
\label{eqn:cascade-model}
\end{equation}
This model is quite restrictive since it allows at most one click per query instance.

The {\bf dependent click model} (DCM) \cite{dcm:guoliuwang} generalizes the
cascade model to instances with multiple clicks:
\begin{eqnarray*}
\Pr(E_i = 1 | E_{i-1} =  1, C_{i-1} = 1) & = & \lambda_i, \\
\Pr(E_i = 1 | E_{i-1} =  1, C_{i-1} = 0) & = & 1.
\label{eqn:DCM}
\end{eqnarray*}
The authors suggest estimating the position effects $\lambda_i$ using maximum
likelihood.

The {\bf user browsing model (UBM)} \cite{dupret:piwowarski} is also based on
the examination hypothesis, but unlike the cascade model and DCM, does not force
$\Pr(E_i = 1 | E_{i-1} = 1, C_{i-1} = 0)$ to be 1.  In other words, it allows
users to stop browsing the current results and instead reformulate the query (or
perhaps give up).  UBM assumes that the examination probability is determined by
the preceding click position.

The {\bf Bayesian browsing model} (BBM) \cite{liu:bbm} uses exactly the same
browsing model as UBM.  However, BBM adopts a Bayesian paradigm for
relevance, i.e., BBM considers relevance to be a random variable with a
probability distribution, rather than a fixed (but unknown) value to be
estimated.  In the context of this paper, where we are focused on the user
browsing model, UBM and BBM can be considered equivalent.

\subsection{Examination depends on prior clicks and prior relevance}
\label{sec:background-relevance}

Next, we summarize models that take into account both clicks on prior results,
and the relevance of those results, to predict the probability of examination.

The {\bf click chain model} (CCM) \cite{ccm:guoliukannan} is a generalization of
DCM obtained by parameterizing $\lambda_i$ and by allowing the user to abandon
examination of more results:
\begin{eqnarray*}
\Pr(E_i = 1 | E_{i-1} =  1, C_{i-1} = 0) & = & \alpha_1 \\
\Pr(E_i = 1 | E_{i-1} =  1, C_{i-1} = 1) & = & \alpha_2 (1-r_{\phi(i-1)}) \\
+  \alpha_3 r_{\phi(i-1)}.
\label{eqn:CCM}
\end{eqnarray*}
Thus if a user clicks on the previous result, the probability that they go on to
examine more results
ranges between $\alpha_2$ and $\alpha_3$ depending on the relevance of the previous result.

The {\bf general click model} (GCM) \cite{zhu:gcm} treats all
relevance and examination effects in the model as random variables:
\begin{eqnarray*}
\Pr(E_i = 1 | E_{i-1} =  1, C_{i-1} = 0) & = & \Pi(A_i>0) \\
\Pr(E_i = 1 | E_{i-1} =  1, C_{i-1} = 1) & = & \Pi(B_i>0) \\
\Pr(C_i = 1 | E_{i}) & = & \Pi(r_{\phi(i)}>0).
\label{eqn:GCM}
\end{eqnarray*}
This allows online inference within the cascade family.
These authors show that all previous models are special cases by suitable choice of the
random variables $A_i, B_i$, and $r_{\phi(i)}$.

\subsection{Post-click models}
\label{sec:background-post-click}

In our discussion so far, relevance referred to ``perceived'' relevance --
whether the user considers the result relevant before she
clicks on the result.  Post-click relevance is a measure of whether the user was
satisfied with their experience after clicking on the result.  Perceived
relevance is positively correlated with post-click relevance
\cite{bouncerate:sculley}.  However, there are cases where perceived relevance
is high and post-click relevance is low (e.g., snippet or creative is
inaccurate), or vice versa (e.g., only a small fraction of people searching
``yahoo'' want answers.yahoo.com -- but for those people, it's perfect). Thus
both perceived and post-click relevance are equally important for user satisfaction.

The {\bf dynamic bayesian model} (DBM)
\cite{chapelle09:bayesian_click_model} uses the ``user satisfaction''
(post-click relevance) of the preceding click to predict whether the
user will continue examining additional results:
\begin{eqnarray*}
\Pr(E_i = 1 | E_{i-1} =  1, C_{i-1} = 0) & = & \gamma \\
\Pr(E_i = 1 | E_{i-1} =  1, C_{i-1} = 1) & = & \gamma (1-s_{\phi(i-1)}),
\label{eqn:DBM}
\end{eqnarray*}
where $s_{\phi(i-1)}$ is the satisfaction of the user in the previous
clicked result. They propose an EM-type estimation method to estimate
$\gamma$ and the user satisfaction variables.

The {\bf session utility model} (SUM)~\cite{intrinsic:dupret} proposes
a user browsing model based on the ``intrinsic'' (post-click) relevance of the
sequence of clicked results in a user session.  However, it does not
model examination or pre-click relevance.

The {\bf click yield prediction} framework \cite{dawei:cy} focuses on predicting
group performance in sponsored search, and  incorporates multiple factors,
such as latent bias, textual features, interactive influences, and ad position
correlation.

{\bf Coupled logistic regression} (CLR) \cite{ning:clr} uses all features from ad,
user, context and nonlinear features such as conjunction information.
It integrates the conjunction information by employing factorization machine
to achieve precise CTR prediction result.

Borisov et al. \cite{alexey:nc} introduces a distributed representation-based
approach for modeling user behavior and several neural click models,
which learn patterns of user behavior directly from interaction data.
{\bf Click sequence model} (CSM) \cite{alexey:csm} predicts the order in which
a user will interact with search engine results. CSM is based on neural network
which follows the encoder-decoder architecture, and it uses an attention mechanism
to extract necessary information about the search engine results.

Zhao et al. \cite{qian:gaze} predicts gaze by combining user browsing data
with eye tracking data from a small number of users in a grid-based recommender
interface. They use Hidden Markov Models (HMMs), and evaluate their prediction
models in MovieLens.

Our focus in this paper is on correctly estimating examination and perceived
relevance.  Thus in the rest of the paper, we will use ``relevance''
as shorthand for ``perceived relevance''.

\section{Micro-Browsing Model}
\label{sec:model}
In this section, we describe the proposed micro-browsing model for
search snippets.

\subsection{Model}
\label{subsec:micro-model}

For a given query $q$, let $r_i$ be the probability that the term in
the $i^{th}$ position in a snippet $R$ is relevant to the query, and
$v_i$ indicate whether that term has been examined by the user
issuing that query, where $r_i \in [0,1]$ and $v_i \in
\{0,1\}$. The probability of relevance of $R$ to $q$ is then given by:
\begin{equation}
\Pr(R|q) = \prod_{i=1}^{m} r_i^{v_i},
\label{eq:prob}
\end{equation}
where $m$ is the number of terms in the snippet $R$. Note that for ease
of notation, we consider here that the snippet $R$ consists of one line
having $m$ terms --- however this analysis can be easily generalized
to a snippet with $m$ lines, in which case the variables $r$ and $v$
will be indexed by both the line and position numbers.

Note that this model considers that not all terms in a snippet are
examined by the user --- only a subset of terms (with corresponding
$v=1$) are examined, and the relevance of the snippet is judged by
the user based on the relevance of only these observed terms.

\subsection{Implications for Snippet CTR}

Let us consider another snippet $S$, for which the relevance and
examination probabilities are indicated by $s$ and $w$ respectively.
Given these two snippets $R$ and $S$ for the query $q$, the ratio of
the probabilities of $R$ and $S$ based on Equation~\ref{eq:prob} is:
\begin{equation}
\frac{\Pr(R|q)}{\Pr(S|q)} = \frac{\prod_{i=1}^{m} r_i^{v_i}}
                                 {\prod_{j=1}^{n} s_j^{w_j}}
\label{eq:ratio}
\end{equation}

One way to empirically validate the effectiveness of our
micro-browsing model for search snippets is to use a dataset where
multiple search snippets relevant to a particular query have different
CTRs --- the effectiveness of the micro-browsing model can then be
measured by it's accuracy in predicting which snippet has a higher
CTR. As we will describe in Section~\ref{sec:data}, one such dataset
is available in sponsored search, comprising of pairs of snippets that
have differences in their text and one where snippet has significantly
higher CTR than another. We use the micro-browsing model to derive
features that are used in training a {\em snippet classifier} ---
given a pair of snippets, the classifier is trained to predict which
snippet has higher CTR.

Let us consider a classifier that, for a given query $q$ and a pair of
creatives $R$ and $S$, predicts which creative is more relevant for
$q$. This classifier takes features from $R$ and $S$ as input, and
outputs a 0/1 prediction.

We derive the features of this classifier from the micro-browsing
model of search results, outlined in
Section~\ref{subsec:micro-model}. The log of the probability ratio in
Equation~\ref{eq:ratio} gives us a score indicative of the relative
relevance of the two snippets given the query $q$:
\begin{equation}
score(R \rightarrow S|q) = \sum_{i=1}^{m} v_i \log r_i - \sum_{j=1}^{n} w_j \log s_j
\label{eq:logratio}
\end{equation}

This motivates us to consider rewrite features that rewrite terms in
$R$ to terms in $S$ in the classifier, details of which are discussed
in the next section.

\section{Predicting Which Snippet Will Have Higher CTR}
\label{sec:technical}
The snippet classifier is trained using the following features.

\subsection{Snippet Classifier Features}

\noindent {\bf Term Features:} The snippet is tokenized to get terms ---
unigrams, bigrams, and trigrams --- from its text. The position of a
term in a line and the number of the line in the snippet are also
considered as features.\\

\noindent {\bf Rewrite Features:} These features are extracted from pairs of
snippets that are given as input to the snippet classifier.  Consider
the following example of two snippets:

\begin{itemize}
\item Snippet 1 ($R$): \\
    line1: ``XYZ Airlines" \\
    line2: ``\textbf{Find cheap} {\it flights} to New York." \\
    line3: ``No reservation costs. Great rates" \\
\item Snippet 2 ($S$): \\
    line1: ``XYZ Airlines" \\
    line2: ``{\it Flying} to New York? \textbf{Get discounts}." \\
    line3: ``No reservation costs. Great rates!"
\end{itemize}

In this pair of snippets, one rewrite (terms bolded in the example) is
from the term `find cheap' in position 1 in line 2 of Snippet 1 ($R$),
to the term `get discounts' in position 5 in line 2 of Snippet 2
($S$). This gives us the rewrite tuple (find cheap:1:2, get
discounts:5:2). Note that there are other rewrites (terms italicized
in the example), e.g., ``flights'' in $R$ got rewritten to ``flying''
in $S$. So, overall the terms (``find cheap'', ``flights'') in $R$ got
rewritten to the terms (``flying'', ``get discounts'') in $S$. Finding
out which phrase in $R$ matches to which corresponding phrase in $S$
is a combinatorial problem in general. In this example, the best match
is ``find cheap'' $\rightarrow$ ``get discounts'' and ``flights''
$\rightarrow$ ``flying'' --- we estimate the most likely matches using
statistics available from our snippet corpus.

The rewrite matching algorithm we use is as follows: given a pair of
snippets differing in one particular phrase rewrite, we assign a score
to that phrase rewrite based on how much improvement (lift) in
observed click-through rate we observe between the two snippets
(details in Section~\ref{sec:experiments}). Collecting these scores
across all possible snippet pairs in the complete snippet corpus gives
us a database of phrase rewrites with corresponding click-through rate
lift scores. To find the matching phrases between $R$ and $S$ in a
rewrite, we greedily match terms in $R$ with corresponding terms in
$S$ that have a high score in the rewrite database --- the intuition
here is that a more probable rewrite like ``find cheap'' $\rightarrow$
``get discounts'' has a higher score in the rewrite database than a
less probable rewrite like ``find cheap'' $\rightarrow$ ``flying''.

After this matching, there could be additional terms in $R$ (not in
$S$), or terms in $S$ (not in $R$) --- we add them as individual
term-level features.

Note that these term feature and the rewrite features can be derived
from the micro-browsing model in Section~\ref{sec:model}. Let us
assume that $S$ can be created from $R$ by rewriting a subset of
features in $R$ to correspoding features in $S$ --- the positions of
the rewrites are encoded in a set of pairs $pair(R,S)$, where
$(p,q) \in pair(R,S)$ indicates that a feature in position $p$ of $R$
has been rewritten to a feature in position $q$ of $S$. We also
consider the set of positions in $R$ and $S$ that occur in $pair(R,S)$
to be $pos(R)$ and $pos(S)$ respectively. Using $pair(R,S)$, $pos(R)$
and $pos(S)$, we can re-factor Equation~\ref{eq:logratio} to the
following form:

\begin{eqnarray}
&&score(R \rightarrow S|q) = \sum_{(p,q) \in pair(R,S)} (v_p \log r_p - w_q \log s_q) \nonumber \\
            && + \sum_{a \notin pos(R)} v_a \log r_a - \sum_{b \notin pos(S)} w_b \log s_b
\label{eq:matched}
\end{eqnarray}

In the example above, $pos(R) = \{1, 3\}$, $pos(S) = \{1, 5\}$, and
$pair(R,S) = \{\{1,5\}, \{3,1\}\}$.

The first term in Equation~\ref{eq:matched} corresponds to the rewrite
features, where the feature in position $p$ in $R$ is rewritten to the
feature in position $q$ in $Q$ -- $v_p$ indicates whether the user saw
the feature in position $p$ in result $R$, while $\log r_p$ gives the
relevance of the feature in position $p$ in result $R$ (similarly for
$S$). The second term enumerates over any terms that are in $R$ but
not in $S$ after the matching, and the third term enumerates over any
terms in $S$ that are in $R$ after the matching. The matching in
$pair(R,S)$ can be done by various mechanisms --- as explained above,
we follow a greedy algorithm to find the most likely match based on
statistics computed from the data source.

\subsection{Snippet Classification Framework}

\begin{figure}
\centering
\includegraphics[scale=0.38]{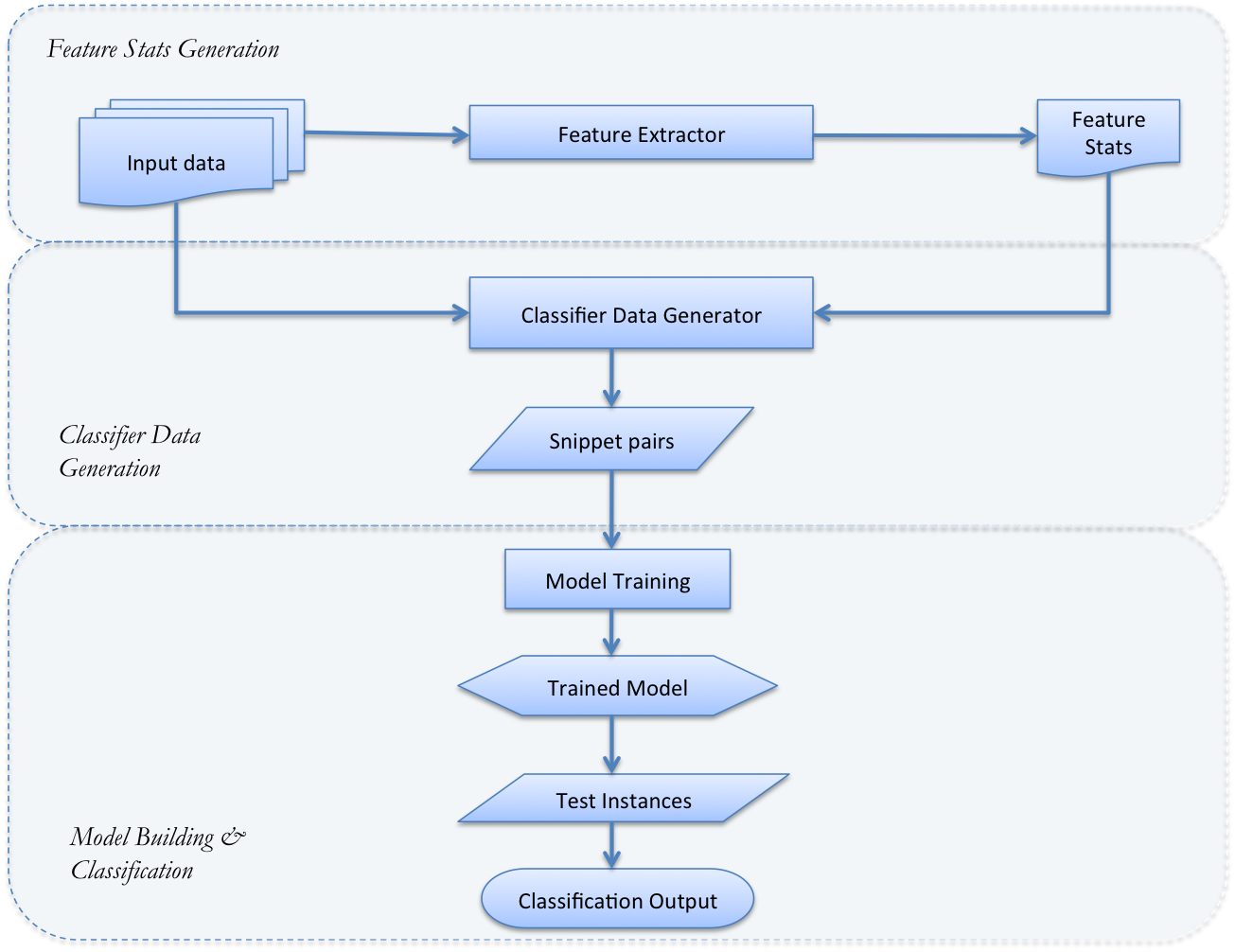}
\caption{Snippet Classification Framework.}
\label{fig:framework}
\end{figure}

The input data for the snippet classifier consists of pairs of
snippets with CTR information for each snippet. The snippet
classification pipeline consists of two phases
(Figure~\ref{fig:framework} shows the overall flow).

In the first phase, the feature extractor extracts important feature
statistics (e.g., rewrites database) from the snippet pairs.  In the
second, the snippet classifier is trained. Each data instance input to
the second phase is a snippet pair $(R, S)$, and the classifier
predicts whether $R$ is better than $S$ or vice versa.  The classifier
data generator takes the snippet corpus and feature statistics
database, and then for all possible snippet pairs it generates the
input features for the classifier. After the model has been trained,
we use it for classifying test data.

\section{Experimental Results}
\label{sec:experiments}
\makeatletter
\renewcommand*{\thetable}{\arabic{table}}
\renewcommand*{\thefigure}{\arabic{figure}}
\let\c@table\c@figure
\makeatother

We conducted a set of experiments to evaluate the effectiveness of the
snippet classifier derived from the micro-browsing model.  As
mentioned in Section~\ref{sec:model}, we used a dataset from sponsored
search for our evaluation.

Before we give the details of the dataset and the experiments, let us
introduce some terminology.  A sponsored search result (or ad)
consists of {\em creative} text that is a brief (e.g., 3 line)
description to be displayed by the search engine, and a {\em keyword}
that specifies the query for which the creative is eligible for
display. An advertiser often groups creatives of the same type (e.g.,
targeting a particular keyword) into a group called the {\em
adgroup}. An ad {\em impression} results when the search engine
displays an ad in response to a user query, and a {\em clickthrough}
results when the user viewing the ad clicks on it and visits the
landing page of the ad. The ad CTR for a given query measures the
fraction of times an ad impression led to a clickthrough.

\subsection{Data Set}
\label{sec:data}

Advertisers often provide multiple alternative creative texts in a
particular adgroup, corresponding to the same keyword used for
targeting queries. When these creatives are shown corresponding to a
query and the keyword used for targeting is the same, any observed
difference in CTR can only caused by difference is the text of the
creative.

We collected such a dataset of creative pairs from an advertiser,
where the keyword used for targeting was same and the observed CTR was
different. Our {\sc AdCorpus} consists of ad creatives collected from
a particular time period, where each adgroup got at least one
click in that time. {\sc AdCorpus} contains tens of millions
creatives pairs, collected from several million adgroups.

\subsection{Serve Weight}

The serve-weight ({\em sw}) of a creative in an adgroup denotes the
probability that the creative will be shown from the set of creatives
of an adgroup. It is computed from clicks and impressions of the
different creatives in the adgroup, suitably normalized by the average
CTR of the adgroup --- this allows serve-weight values of two
creatives in different adgroups to be compared, as it accounts for the
CTR differences between adgroups.

In Section~\ref{sec:technical}, we described how term and rewrite
features were generated from a pair of creatives.  We compute the {\em
  sw-diff} for each term feature in a creative pair --- if the term is
present in one creative but not the other, we define sw-diff as the
difference in serve-weight of the creative containing that term with
the serve-weight of the creative not containing that term.  For
rewrite features, if a term in creative $R$ is rewritten to a term in
creative $S$ for a creative pair, we define sw-diff as the difference
of serve-weights of $R$ and $S$.  For both term and rewrite features,
we define a random variable {\em delta-sw} that is set to +1 if
sw-diff is positive for that feature and -1 if sw-diff is negative.

\subsection{Feature Statistics Database}
\label{sec:database}

For each feature, we compute the empirical probability $p$ of sw-diff
being +1 by estimating the fraction of times delta-sw is +1 over the
complete {\sc AdCorpus} (using Laplace-smoothing to address
sparsity). We record the odds-ratio of this probability
($\frac{p}{1-p}$) as the statistic corresponding to that feature in
the statistics database.  The statistics for each feature in this
database intuitively estimates the odds of the presence of the feature
causing an increase in creative CTR (since it increases
serve-weight). Note that apart from term and rewrite features, we also
calculate position features, i.e., for positions of terms and position
pairs (source position and target position) for rewrites, we estimate
the empirical probability of sw-diff being +1.

\subsection{Experiments}
We conducted a set of experiments where we systematically ablated
features from the classifier model to estimate the effectiveness of
different aspects of the snippet classifier derived from the
micro-browsing model. Specifically, we consider the following
components in our ablation study:

\begin{itemize}

\item Term features: Use n-gram (unigram, bigram, trigram) features from the input creative pair in the classifier. This is equivalent to only using the second and the third terms in Equation~\ref{eq:matched} in the score for the micro-browsing model, where $v_a$ and $w_b$ are set to 1 for all terms (i.e., all terms are viewed).

\item Rewrite features: Use term rewrite features for an input creative pair in the classifier.  This is equivalent to only using the first term in Equation~\ref{eq:matched} in the score for the micro-browsing model, where $v_a$ and $w_b$ are set to 1 for all terms (i.e., all terms are viewed).

\item Position information: Uses position information of terms in the term and rewrite features in the classifier. This is equivalent to considering $v_a$ and $w_b$ in Equation~\ref{eq:matched}, that can now model which terms are viewed by the user.

\item Feature statistics database: Use the feature statistics estimated from {\sc AdCorpus} as the initial values of the term and rewrite features in the learning algorithm.

\end{itemize}

The snippet classifier we train is a logistic regression model with L1
regularization.  We turn on these individual components incrementally
in the feature set of the logistic regression model, to create
multiple snippet classifier models:

\begin{itemize}
\item
M1: Term features with no position information, where feature values are initialized
from the feature statistics database.
\item
M2: Term features with position information, where feature values are initialized
from the feature statistics database.
\item
M3: Greedy rewrite features with no position information, where feature values are initialized
from the feature statistics database.
\item
M4: Greedy rewrite features with position information, feature values initialized using the feature statistics database.
\item
M5: Greedy rewrite and term features with no position information, where feature values are initialized
from the feature statistics database.
\item
M6: Greedy rewrite and term features with position information, feature values initialized using the feature statistics database.
\end{itemize}

\subsubsection{Mapping of Micro-browsing model to Classifier}

Let us now consider the mapping of the features from the
micro-browsing model to the snippet classifier for these 5
variants. We illustrate this using M4, which considers rewrite
features with position information and initialization. Let us consider
only the rewrite features in Equation~\ref{eq:matched}, which is
derived from the micro-browsing model:
\begin{equation}
score(R \rightarrow S|q) = \sum_{(p,q) \in pair(R,S)} (v_p \log r_p - w_q \log s_q)
\label{eq:rewrite}
\end{equation}

In the logistic regression model, whenever we see a rewrite feature
for a pair of creatives $R$ and $S$, we lookup the feature statistics
database to see if we have any feature statistics corresponding to
that rewrite and initialize the logistic regression learning algorithm
with those feature values. As described in
Section~\ref{sec:technical}, we collect feature statistics for
rewrites independent of position of the rewrite terms, to handle
sparsity issues (since if we consider term x position information in
the rewrites, the data would be too sparse).  This corresponds to
making the following approximation to Equation~\ref{eq:rewrite}:
\begin{equation}
score(R \rightarrow S|q) = \sum_{(p,q) \in pair(R,S)} f(v_p, w_q) \log (r_p/s_q)
\label{eq:rewrite-approx}
\end{equation}
That is, the position terms and the relevance terms are decoupled, so
that the relevance terms can be initialized using the features
statistics database. Note that we also initialize the position features
$f(v_p, w_q)$ using the rewrite position features
calculated from {\sc AdCorpus}, as described in Section~\ref{sec:database}.

Based on Equation~\ref{eq:rewrite-approx}, we represent the logistic
regression problem for M4 as:
\begin{equation}
\log O = \sum_{(p,q) \in pair(R,S)} P_{p,q} T_{p,q},
\label{eq:m4}
\end{equation}
where O is the odds of $R$ having higher CTR than $S$, $P$ are the
classifier features corresponding to positions of the rewrite terms,
and $T$ are the classifier features corresponding to the relevance of
the rewrite terms. If we fix the values of $P$, $T$ can be learned as
a logistic regression model. Similarly if we fix the values of $T$,
$P$ can be learned as a logistic regression model. So, learning model
M4 can be framed as an iterative learning of features $P$ and $T$ in
Equation~\ref{eq:m4} using two coupled logistic regression
models. Note that it can be shown that the iterative learning of $P$
and $T$ in Equation~\ref{eq:m4} converges to a fixed point under
certain model assumptions, but that is beyond the scope of this paper.

\begin{table}
\centering
\caption{Accuracy of creative classification using different sets of features}
\begin{tabular}{|c|c|c|c|} \hline
Feature & Recall & Precision & F-Measure \\ \hline
M1: Terms only & 55.9\% & 58.2\%  & 0.570 \\ \hline
M2: Terms w. pos & 64.4\% & 66.3\%  & 0.653 \\ \hline
M3: Rewrites only & 59.0\% & 61.2\%  & 0.601 \\ \hline
M4: Rewrites w. pos & 70.0\% & 71.9\%  & 0.709 \\ \hline
M5: Rewrites \& terms & 59.7\% & 61.8\%  & 0.607 \\ \hline
M6: Rewrites \& terms w. pos & 70.4\% & 72.1\%  & 0.712 \\ \hline
\end{tabular}
\label{table:ablation}
\end{table}

\subsubsection{Analysis of Results}

We performed standard 10-fold cross validation experiments,
where in each cross validation iteration 90\% instances are used for
training and the rest 10\% are used for
testing. Table~\ref{table:ablation} shows the precision, recall and
F-measure of creative classification for these different models. As
seen from the table, the baseline model is M1, which only uses terms
from the creatives in the classifier, as done in a usual bag-of-terms
model. As we add each component, based on features derived from the
micro-browsing model, the classifier performance keeps improving. The
final model M6, which incorporates all aspects of the micro-browsing
model, has a dramatic improvement in performance compared to M1 ---
the F-measure increase from $0.57$ for M1 to $0.71$ for M6.

\begin{figure}
\centering
\includegraphics[scale=0.40]{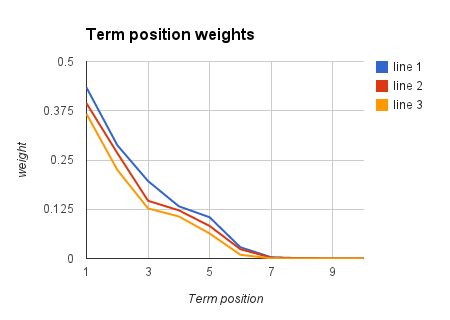}
\caption{Learned term position weights (line 1,2,3).}
\label{fig:learned-term-weights}
\end{figure}

Let us now look at the values of the position features corresponding
to the terms and rewrites.  Figure~\ref{fig:learned-term-weights}
shows the learned term position weights for line 1, 2 and 3.

Table~\ref{table:toprhs} presents the accuracy of creative classification for top and right-hand-side(rhs) ads.
The accuracy of the classifier using the top ads data is slightly higher than that of rhs data.

\begin{table}
\centering
\caption{Accuracy of creative classification in different configuration (top vs. rhs) }
\begin{tabular}{|c|c|c|} \hline
Feature & Top & Rhs \\ \hline
M1: Terms only & 57.1\%  & 57.0\% \\ \hline
M2: Terms w. position & 65.7\%  & 65.1\% \\ \hline
M3: Rewrites only & 60.2\%  & 59.9\% \\ \hline
M4: Rewrites w. position & 71.1\%  & 70.8\% \\ \hline
M5: Rewrites and terms & 60.9\%  & 60.6\% \\ \hline
M6: Rewrites and terms w. position & 71.4\%  & 71.1\% \\ \hline
\end{tabular}
\label{table:toprhs}
\end{table}

\section{Conclusions and Future Work}
\label{sec:conclusions}
In this paper, we proposed a {\em micro-browsing model} for search result snippets,
and presented an application of this model to predict better creative.
We discovered that the micro-position of the terms as well as the rewrites is an
important feature and influences the accuracy of the classifier.

Future work directions include learning the micro-position normalizers and
adding more features to improve the accuracy of the classifier.
Another promising direction will be to use language models to have deeper
understanding of snippet text, as well as automatic generation of snippets.
We would also like to do eye-tracking \cite{qian:gaze}
studies to see how the positions of important words in the snippet correlates
with focus areas identified by the eye tracking models. Finally, we would like
to study how our approach can be integrated with attention-based neural
network models (e.g., transformer networks \cite{vaswani:attn}).


\begin{thebibliography}{10}

\bibitem{agichtein06:improve_web_search}
E.~Agichtein, E.~Brill, and S.~T. Dumais.
\newblock Improving web search ranking by incorporating user behavior.
\newblock In {\em SIGIR}, 2006.

\bibitem{alexey:nc}
A.~Borisov, I.~Markov, M.~de~Rijke, and P.~Serdyukov.
\newblock A neural click model for web search.
\newblock In {\em WWW}, 2016.

\bibitem{alexey:csm}
A.~Borisov, M.~Wardenaar, I.~Markov, and M.~de~Rijke.
\newblock A click sequence model for web search.
\newblock In {\em SIGIR}, 2018.

\bibitem{chapelle09:bayesian_click_model}
O.~Chapelle and Y.~Zhang.
\newblock A dynamic bayesian network click model for web search ranking.
\newblock In {\em WWW}, 2009.

\bibitem{chen12:noise_aware}
W.~Chen, D.~Wang, Y.~Zhang, Z.~Chen, A.~Singla, and Q.~Yang.
\newblock A noise-aware click model for web search.
\newblock In {\em WSDM}, 2012.

\bibitem{dcm:craswell}
N.~Craswell, B.~Ramsey, M.~Taylor, and O.~Zoeter.
\newblock An experimental comparison of click position-bias models.
\newblock In {\em WSDM}, 2008.

\bibitem{intrinsic:dupret}
G.~Dupret and C.~Liao.
\newblock A model to estimate intrinsic document relevance from the
  clickthrough logs of a web search engine.
\newblock In {\em WSDM}, 2010.

\bibitem{dupret:piwowarski}
G.~Dupret and B.~Piwowarski.
\newblock A user browsing model to predict search engine click data from past
  observations.
\newblock In {\em SIGIR}, 2008.

\bibitem{ccm:guoliukannan}
F.~Guo, C.~Liu, A.~Kannan, T.~Minka, M.~Taylor, Y.~Wang, and C.~Faloutsos.
\newblock Click chain model in web search.
\newblock In {\em WWW}, 2009.

\bibitem{dcm:guoliuwang}
F.~Guo, C.~Liu, and Y.~Wang.
\newblock Efficient multiple-click models in web search.
\newblock In {\em WSDM}, 2009.

\bibitem{joachims02:optimize_search_engines}
T.~Joachims.
\newblock Optimizing search engines using clickthrough data.
\newblock In {\em KDD}, 2002.

\bibitem{liu:bbm}
C.~Liu, F.~Guo, and C.~Faloutsos.
\newblock Bbm: Bayesian browsing model from petabyte-scale data.
\newblock In {\em KDD}, 2009.

\bibitem{radlinski05:query_chains}
F.~Radlinski and T.~Joachims.
\newblock Query chains: Learning to rank from implicit feedback.
\newblock In {\em KDD}, 2005.

\bibitem{predict:richardson}
M.~Richardson, E.~Dominowska, and R.~Ragno.
\newblock Predicting clicks: Estimating the click-through rate for new ads.
\newblock In {\em WWW}, 2007.

\bibitem{bouncerate:sculley}
D.~Sculley, R.~Malkin, S.~Basu, and R.~J. Bayardo.
\newblock Predicting bounce rates in sponsored search advertisements.
\newblock In {\em KDD}, 2009.

\bibitem{srikant:rve}
R.~Srikant, S.~Basu, N.~Wang, and D.~Pregibon.
\newblock User browsing models: relevance versus examination.
\newblock In {\em KDD}, 2010.

\bibitem{vaswani:attn}
A.~Vaswani, N.~Shazeer, N.~Parma, J.~Uszkoreit, L.~Jones, A.~N. Gomez, Łukasz
  Kaiser, and I.~Polosukhin.
\newblock Attention is all you need.
\newblock In {\em NIPS}, 2017.

\bibitem{dawei:cy}
D.~Yin, B.~Cao, J.-T. Sun, and B.~D. Davison.
\newblock Estimating ad group performance in sponsored search.
\newblock In {\em WSDM}, 2014.

\bibitem{ning:clr}
N.~Yin, H.~Li, and H.~Su.
\newblock Clr: coupled logistic regression model for ctr prediction.
\newblock In {\em ACM TUR-C}, 2017.

\bibitem{qian:gaze}
Q.~Zhao, S.~Chang, F.~M. Harper, and J.~A. Konstan.
\newblock Gaze prediction for recommender systems.
\newblock In {\em RecSys}, 2016.

\bibitem{zhu:gcm}
Z.~Zhu, W.~Chen, T.~Minka, C.~Zhu, and Z.~Chen.
\newblock A novel click model and its applications to online advertising.
\newblock In {\em WSDM}, 2010.

\end{thebibliography}
\end{document}